\documentclass[manuscript]{acmart}

\AtBeginDocument{%
  }

\copyrightyear{2024}
\acmYear{2024}
\setcopyright{rightsretained}
\acmConference[ICDSP 2024]{2024 8th International Conference on Digital Signal Processing (ICDSP)}{February 23--25, 2024}{Hangzhou, China}
\acmBooktitle{2024 8th International Conference on Digital Signal Processing (ICDSP) (ICDSP 2024), February 23--25, 2024, Hangzhou, China}\acmDOI{10.1145/3653876.3653899}
\acmISBN{979-8-4007-0902-9/24/02}

\usepackage{amsmath,amsfonts}
\usepackage{algorithmic}
\usepackage{algorithm}
\usepackage{url}
\usepackage{graphicx}

\begin{document}

\title{Deep Reinforcement Learning with~Swin~Transformers}
\author{Li Meng}
\email{li.meng@its.uio.no}
\orcid{0000-0002-8867-9104}
\affiliation{%
  \institution{Department of Technology Systems, University of Oslo}
  \streetaddress{P.O box 70}
  \city{Kjeller}
  \country{Norway}
  \postcode{N-2027}
}

\author{Morten Goodwin}
\orcid{0000-0001-6331-702X}
\affiliation{%
  \institution{Centre for Artificial Intelligence Research, University of Agder}
  \country{Norway}}

\author{Anis Yazidi}
\orcid{0000-0001-7591-1659}
\affiliation{%
  \institution{Oslo Metropolitan University}
  \country{Norway}
}

\author{Paal Engelstad}
\orcid{0009-0000-8371-927X}
\affiliation{%
 \institution{Department of Technology Systems, University of Oslo}
 \country{Norway}}

\renewcommand{\shortauthors}{Meng et al.}

\begin{abstract}
Transformers are neural network models that utilize multiple layers of self-attention heads and have exhibited enormous potential in natural language processing tasks. Meanwhile, there have been efforts to adapt transformers to visual tasks of machine learning, including Vision Transformers and Swin Transformers. Although some researchers use Vision Transformers for reinforcement learning tasks, their experiments remain at a small scale due to the high computational cost. This article presents the first online reinforcement learning scheme that is based on Swin Transformers: Swin DQN. In contrast to existing research, our novel approach demonstrate the superior performance with experiments on 49 games in the Arcade Learning Environment. The results show that our approach achieves significantly higher maximal evaluation scores than the baseline method in 45 of all the 49 games ( \(\sim \) 92 \%), and higher mean evaluation scores than the baseline method in 40 of all the 49 games (\(\sim \) 82 \%). The code can be found on \url{https://github.com/mengli11235/SWIN-DQN}.
\end{abstract}





\received{5 January 2024}
\received[revised]{22 January 2024}
\received[accepted]{22 January 2024}

\maketitle

\section{Introduction}
Reinforcement learning (RL) and natural language processing (NLP) are considered sub-fields of Artificial Intelligence (AI). They both deal with temporal sequences with the possibility of using neural networks (NNs). State-of-the-art solutions to RL rely on the Markov decision process (MDP) to build the temporal decision-making model, which describes a discrete-time stochastic control process that can be solved mathematically through dynamic programming, while transformers are a common solution to handle NLP tasks.


Prior to the introduction of transformers, the most common deep learning (DL) solution to handle NLP tasks was to utilize recurrent neural networks (RNNs). Transformers use a self-attention network and demonstrate more robust capabilities than previous attention-based RNNs \cite{vaswani2017attention}. One main improvement from the Transformer is to encode each hidden state with its attention-based context vector, which yields significantly richer contextual representations when stacked over multiple NN layers. The contextual information is also exempted from sequential strictness. Another main improvement is to use positional embeddings to encode relative positions of the input tokens.

Recent research has demonstrated the strength of convolutional neural networks (CNNs) on real-world applications \cite{li2021image, gao2024image, li2023synthesizing, li2023terrain, li2020exertion, li2022interactive, li2022musical, li2023location, li2023sceneaware}, and focused on the viability of replacing CNNs with transformers in other domains of machine learning (ML). Vision Transformers (ViT) successfully adapted transformers from the NLP domain into the computer vision (CV) field \cite{dosovitskiy2020image}. Swin Transformers further improved ViT in terms of both computational overhead and accuracy \cite{liu2021swin}.


RL studies intelligent behaviors in environments that typically require an agent to maximize the obtained rewards. One of RL's main challenges is to balance the rate of exploration over exploitation in the agent's planning. The outcomes of past exploration need to be stored, and the RL agent then uses those outcomes to predict the expected rewards in future runs. Q-values are values that record such outcomes of actions at a certain state. Q-learning is an RL algorithm that keeps track of Q-values and uses them to update the policy according to the Bellman Optimality Equation (Eq. \ref{eq:q}) \cite{watkins1992q}. In this equation, $\gamma$ is the discounting factor, $Q^*(s,a)$ is the Q-value at state $s$ of action $a$, $r_{t+1}$ is the result value obtained by advancing to state $s_{t+1}$ at time instant $t+1$.

\scriptsize
\begin{equation}
    Q^*(s,a) = E\{r_{t+1}+\gamma \mathrm{max}_{a'}Q^*(s_{t+1},a'|s_t=s,a_t=a)\}
    \label{eq:q}
\end{equation}
\normalsize


Combining RL policies with NNs has also become a common practice in RL. In particular, CNNs have been heavily applied to games with image displays in Deep reinforcement learning tasks (DRL). Deep Q-learning Network (DQN) is a DRL method that stores Q-values in NNs instead and produces the Q-value predictions as the network output. DQN has achieved human, or superhuman-level performances on many Atari games \cite{mnih2013playing}. There were researches using ViT in RL from image pixels \cite{tao2022evaluating, kargar2021vision, kalantari2022improving}. However, training ViT in RL can be costly due to its quadratic complexity relative to the input image size.

This paper introduces the Swin DQN, an online RL scheme with Swin Transformers. This method extends the well adopted Double Q-learning \cite{hasselt2010double} with recently introduced Swin Transformers. The heart of the method is splitting groups of image pixels into small tokenized patches and applying local self-attention operations inside the (shifted) windows of fixed sizes as an extension. 

\section{Related Work}
\noindent \textbf{DQN with ViT} Kalantari et al. \cite{kalantari2022improving} applied ViT to DQN for the purpose of improving sample complexity. To reduce the costs of ViT, they used the Linformer architecture \cite{wang2020self} and discarded all the tokens except value tokens at the end of the encoder. Nonetheless, they could not obtain complete results at 200M frames on the Atari games they had selected due to limited resources.



\noindent \textbf{Self-supervised Vision Transformers} Self-supervised Vision Transformers \cite{chen2021empirical} combine ViT with self-supervised methods such as Momentum Contrast (MoCo) \cite{he2020momentum}. MoCo builds dynamic dictionaries for contrastive learning in pre-training and the learned representations are to be transferred to downstream tasks. It was observed that instability of ViT is a major issue when used together with MoCo. More generally, there is evidence that self-supervised pre-training of ViT provides clearer semantic segmentation information of images than supervised learning \cite{caron2021emerging}.

\noindent \textbf{Decision Transformers} Decision Transformers view RL problems as the same sequence modeling problems as NLP does, which abandon the traditional value function and policy gradient methods in RL and replace them with causally masked transformers \cite{chen2021decision}. One of the major limitations of Decision Transformers is that it only works on offline RL. States, actions and returns must be known prior to the training of the model. Online Decison Transformers (ODT) tackle this issue by integrating offline pretraining and online finetuning in a unified framework \cite{zheng2022online} and achieve state-of-the-art performance on the D4RL benchmark.  


\section{Method}


Double Q-learning \cite{hasselt2010double} is a widely adopted algorithm that improves upon Q-learning. Q-learning tends to overestimate the Q-values because the policy updates on the target values selected by the $\mathrm{max}$ operator, which perpetuates the approximation errors in the direction of overestimation. Overestimation in Q-learning can have undesired effects and lead to suboptimal performances of an agent \cite{thrun1993issues, meng2021expert}. Double Q-learning avoids overestimation by having two separate Q-value functions. Only one of them is updated at once according to either (Eq. \ref{eq:QA}) or (Eq. \ref{eq:QB}). $\alpha(s,a)$ is the learning rate, $a^*$ is $\mathrm{argmax}_aQ_\theta^A(s',a)$, and $b^*$ is $\mathrm{argmax}_aQ_\theta^B(s',a)$ at the next state $s'$. This update rule also indicates that Double Q-learning suffers from underestimation instead of overestimation because the target approximation is a weighted estimate of unbiased expected values, which is lower or equal to the maximum expected values \cite{hasselt2010double}.

\scriptsize
\begin{equation}
    \label{eq:QA}
    Q^A(s,a) = Q^A(s,a) + \alpha(s,a)(r+\gamma Q^B(s',a^*)-Q^A(s,a))
\end{equation}
\normalsize
\scriptsize
\begin{equation}
    \label{eq:QB}
    Q^B(s,a) = Q^B(s,a) + \alpha(s,a)(r+\gamma Q^A(s',b^*)-Q^B(s,a))
\end{equation}
\normalsize


Algorithm \textbf{\ref{algo:mine}} shows the pseudo-code of our Double DQN. Here, $\gamma$ is the discounting factor, $\epsilon$ is the exploration ratio, $maxFrames$ is the maximal number of total frames, and $L()$ is the loss function, i.e., Smooth L1 loss in this case.

$Q_\theta^A$ stands for the policy network and $Q_\theta^B$ the target network, which is a copy of $Q_\theta^A$. $syncFrames$ is the number of frames between the synchronizations of two networks. $s$ is the current state, $s'$ the next state, $a$ the action, $r$ the reward, and $terminal$ is a flag indicating if the game terminates or not. In Algorithm \textbf{\ref{algo:mine2}}, $a^*$ is $\mathrm{argmax}_aQ_{\theta}^A(s',a)$.

\begin{algorithm}[tb]
\scriptsize
\caption{Double Q-learning}\label{algo:mine}
\begin{algorithmic}[1]
\STATE \textbf{Input}: $\epsilon$, $\gamma$, $maxFrames$, $syncFrames$, $L()$\\
\STATE \textbf{Parameter}: D, $Q_\theta^A$,$Q_\theta^B$\\
\STATE \textbf{Output}: $Q_\theta^A$
\STATE $frames \gets 0$
\WHILE{$frames<maxFrames$}
\STATE Initialize the environment

\WHILE{Game not finished}
\STATE $frames$ += $4$
\IF{$random()<\epsilon$}
\STATE Select an action randomly
\ELSE
\STATE Select an action through $Q_{\theta}^A$
\ENDIF
\STATE Store $s,a,s', r, terminal$ into replay buffer D
\STATE Draw minibatch of $s,a,s', r, terminal$ from D
\STATE Update $Q_\theta^A$ according to Algorithm \ref{algo:mine2}
\ENDWHILE
\IF {$frames\%syncFrames==0$}
\STATE $Q_\theta^B \gets Q_\theta^A$
\ENDIF
\ENDWHILE
\STATE \textbf{return} $Q_\theta^A$
\end{algorithmic}
\end{algorithm}
\normalsize

\begin{algorithm}[tb]
\scriptsize
\caption{Network Update}\label{algo:mine2}
\begin{algorithmic}[1]
\STATE \textbf{Input}: $s,a,s', r, terminal$, $L()$\\
\STATE \textbf{Parameter}: $Q_\theta^A$,$Q_\theta^B$\\
\STATE \textbf{Output}:
\STATE $target = r+\gamma Q^B_{\theta}(s',a^*)*(1-terminal)$
\STATE Compute $loss$ by $L(Q^A_{\theta}(s,a), target)$
\STATE Update $Q_\theta^A$ by $loss$ with gradient descent
\STATE \textbf{return}
\end{algorithmic}
\end{algorithm}
\normalsize

Attention is implemented in transformers as the contextual embeddings of the 'query' ($Q$), 'key' ($K$) and 'value' ($V$), as in (Eq. \ref{eq:attention}). $B$ is the relative position bias that improves over absolute position embedding and $d_k$ is the dimension of the query and key. Transformers allow the re-combination of attention information from different layers, and the processing of all inputs at once, which are more convenient than RNNs when dealing with a large number of data.

\scriptsize
\begin{equation}
    \mathrm{Attention}(Q,K,V) = \mathrm{SoftMax}(QK^T/\sqrt{d_k}+B)V
    \label{eq:attention}
\end{equation}
\normalsize

We improve DQN by replacing CNNs with an adaptation of Swin Transformers, Swin MLP, which implements multi-head self-attention operations as grouped 1D convolutions. We name this new method as Swin DQN. Fig. \ref{fig:swin} shows the architecture. Image pixels are grouped into small patches by one layer of 2D convolutions so that the output channels become hidden embeddings. Each basic layer contains a number of Swin blocks, illustrated on the bottom left in Fig. \ref{fig:swin}. Patches are grouped into local windows prior to the self-attention operation, which is done through a grouped 1D convolutional layer. This exploitation of locality reduces the computational complexity from quadratic to linear when the window size is fixed. Afterwards, there are two linear layers with the number of hidden units proportional to the embedding dimensions, followed by reshaping operations that reverse the window partition.

One shortcoming of utilizing locality is the lack of self-attentions among different local windows. To effectively overcome this, the window partition is shifted in successive blocks to introduce cross-window connections. Namely, the first and third blocks displace the windows by a number of patches, so that the windows are overlapped over different neighboring blocks. Neighboring patches are merged after all but the last basic layers in order to build hierarchical feature maps. Patch merging also reduces the dimensions of embeddings by half by adding an additional linear layer.

\section{Experimental Details}

\begin{figure}[t]
    \centering
    \includegraphics[width=0.4\linewidth]{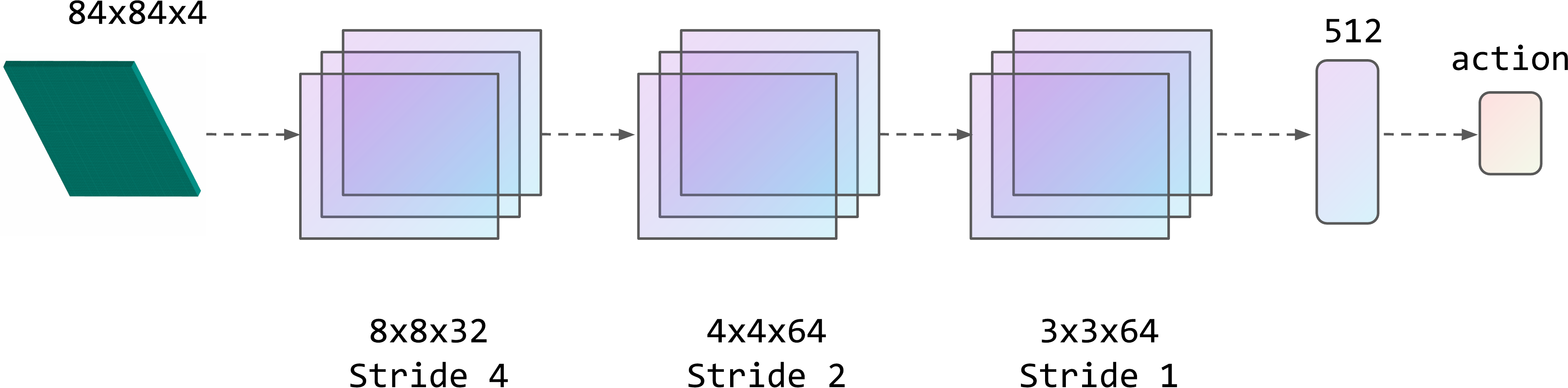}
    \caption{The structure of our DQN. It consists of three convolutional layers, and a fully connected layer, followed by an output layer.}
\label{fig:nn}
\end{figure}

Our experiment is conducted among 49 Atari games of Arcade Learning Environment (ALE). Performances are evaluated based on both the highest evaluation scores and mean evaluation scores. Fig. \ref{fig:nn} shows the NN architecture of our Double DQN. The list of chosen parameters for Double DQN is available in Table \ref{tbl:para}. We use Adam \cite{kingma2014adam} as our optimizer and the learning rate remains the same for both Double DQN and Swin DQN. 

\begin{figure*}[t]
    \centering
    \includegraphics[width=0.6\linewidth]{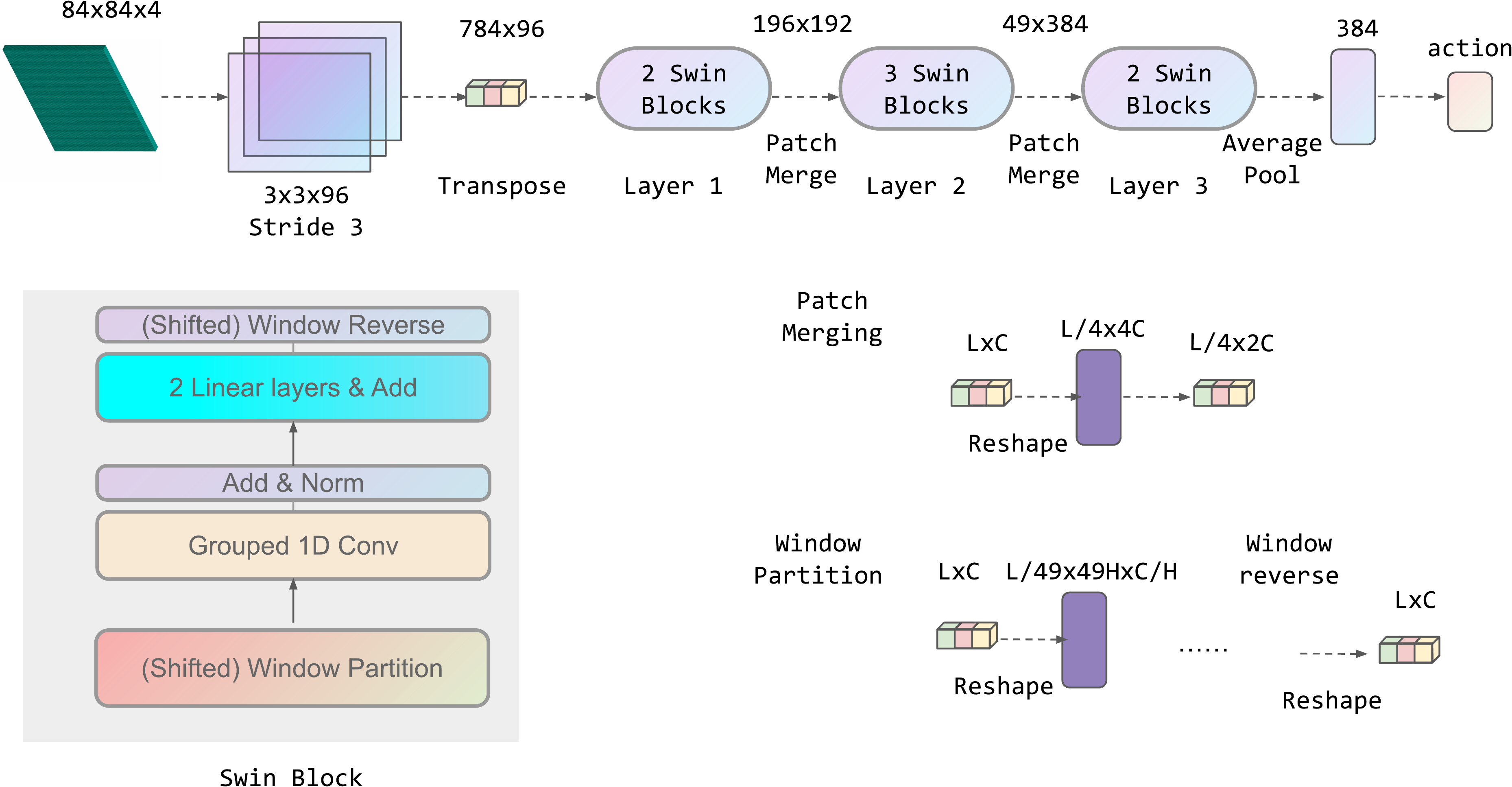}
    \caption{The architecture of our Swin DQN. The top shows the step-by-step procedure. The bottom left box contains structures inside a Swin block. The details of patch merging, window partition and window merging are illustrated on the bottom right.}
\label{fig:swin}
\end{figure*}

Fig. \ref{fig:swin} shows a full illustration of our Swin DQN. The list of parameters specified for Swin DQN is included in Table \ref{tbl:spara}. The rest of the hyper-parameters are kept the same as in Double DQN. We use three layers of Swin blocks in our NN. Those layers contain 2, 3, 2 Swin blocks and 3, 3, 6 attention heads, respectively. The patch size is set to $3\times3$, which yields $28\times28$ patches since the input size is $84\times84$ for each channel. The embedding dimension for each patch is 96. This suggests that the token size after patch embedding is $784\times96$. The matrix operation of self-attentions is conducted through a grouped 1D convolutional layer, where the number of groups is equal to the number of attention heads. The drop path rate $0.1$ states that there is a 10\% chance that the input is kept as it is in skip connections.

There are random no-operation (no-op) steps of $[0,30)$ at the start of each game to help introduce stochasticity into the environment. We set the seed of the Atari environment fixed for the purpose of training, but randomly select seeds for a more concrete evaluation phase. Randomly seeded games are initialized for every single episode of the evaluation.

\begin{table}[t]
\scriptsize
\caption{Parameters used in Double DQN}
\centering
\begin{tabular}{ |c||c|} 
 \hline
    Input & $84\times84\times4$ \\ 
  \hline
      Optimizer & Adam \\ 
  \hline
        Adam learning rate & 0.0000625 \\ 
  \hline
      $\gamma$ & 0.99 \\ 
  \hline
      Initial $\epsilon$ & 1 \\
  \hline
      Final $\epsilon$ & 0.01 \\
  \hline
      $\epsilon$ decay frames & 1M \\
  \hline
      SyncFrames & 40000 \\
    \hline
      Frames per step & 4 \\ 
  \hline
      Steps per update & 4 \\ 
  \hline
      Steps per evaluation & 250000 \\ 
  \hline
      MaxFrames & 200M \\ 
  \hline
      Replay size & 1M\\
  \hline
      Batch size & 32\\
  \hline
\end{tabular}
\label{tbl:para}
\end{table}

\begin{table}[t]
\scriptsize
\caption{Parameters specified for Swin DQN}
\centering
\begin{tabular}{ |c||c|} 
 \hline
    Layers & 3 \\ 
  \hline
      Blocks each layer & 2, 3, 2 \\ 
  \hline
        Heads each layer & 3, 3, 6 \\ 
  \hline
      Patch size & $3\times3$ \\ 
  \hline
      Window size & $7\times7$ \\
  \hline
      Embedding dimension & 96 \\
  \hline
      MLP ratio & 4 \\
  \hline
      Drop path rate & 0.1 \\
  \hline
\end{tabular}
\label{tbl:spara}
\end{table}

\section{Results}


\begin{table*}[t]
\scriptsize
\caption{Maximal Evaluation Scores}
\centering
\begin{tabular}{ |c||c|c c|c c|c|} 
 \hline
    \textbf{Game} & \textbf{Boot-DQN} & \textbf{Double DQN} &\textbf{(Normalized)} &\textbf{Swin DQN} &\textbf{(Normalized)} & \textbf{Nature}\\
  \hline
   Alien & 2436.6 & 5820 &(0.81) & \textbf{16670}&(2.38) & 3069\\
   \hline
    Amidar & 1272.5 & 1421 &(0.83)& \textbf{2665} &(1.55) & 739.5\\
   \hline
    Assault & 8047.1 & 4130&(7.52) & \textbf{8582} &(16.09) & 3359\\
    \hline
    Asterix & 19713.2 & 42900 &(5.15)& \textbf{981500} &(118.32) &  6012\\
    \hline
    Asteroids & 1032 & 2100&(0.03) & \textbf{11360}&(0.23) &1629\\
    \hline
    Atlantis & 994500 & \textbf{1012300}&(61.78) & 1006000&(61.39) & 85641\\
    \hline
    Bank Heist & 1208 & \textbf{1260}&(1.69) & 1230&(1.65) & 429.7\\
    \hline
    Battle Zone & 38666.7 & 57000&(1.57) & \textbf{149000}&(4.21) &  26300\\
    \hline
    Beam Rider & 23429.8 & 18150&(1.07) & \textbf{60936}&(3.66) &  6846\\
    \hline
    Bowling & 60.2 & 30 &(0.05)& \textbf{125}&(0.74)  & 42.4\\
    \hline
    Boxing & 93.2 & \textbf{100} &(8.33) & \textbf{100} &(8.33) & 71.8\\
    \hline
    Breakout & 855 & 431 &(14.91) &\textbf{864} &(29.94) & 401.2\\
     \hline
    Centipede & 4553.5 & 14297&(1.23) & \textbf{23448}&(2.15) & 8309\\
    \hline
    Chopper Command & 4100 & 3400&(0.39) & \textbf{7200} &(0.97) & 6687\\
    \hline
    Crazy Climber & 137925.9 & 218700&(8.3) & \textbf{246900}&(9.43) & 114103\\
    \hline
    Demon Attack & 82610 & 28225 &(15.43) & \textbf{134465}&(73.84) & 9711\\
    \hline
    Double Dunk & \textbf{3} & 2&(9.36) & 2 &(9.36)&  -18.1\\
    \hline
    Enduro & 1591 & 2265 &(2.63)& \textbf{2361} &(2.74) & 301.8\\
    \hline
    Fishing Derby & 26 & 29 &(2.28) & \textbf{43} &(2.54) & -0.8\\
    \hline
    Freeway & 33.9 & 26 &(0.88)& \textbf{34} &(1.15) & 30.3\\
    \hline
    Frostbite & 2181.4 & 3770 &(0.87) & \textbf{6140}&(1.42) & 328.3\\
    \hline
    Gopher & 17438.4 & 83360&(38.56) & \textbf{108640}&(50.3) & 8520\\
    \hline
    Gravitar & 286.1 & \textbf{1100}&(0.29) & 950&(0.24) & 306.7\\
    \hline
    Hero & 21021.3 & 13885&(0.43) & \textbf{31475} &(1.02)& 19950\\
    \hline
    Ice Hockey & -1.3 & 3 &(1.17) & \textbf{5} &(1.34) & -1.6\\
    \hline
    Jamesbond & 1663.5 & 5700&(20.71) & \textbf{9850} &(35.87) & 576.7\\
    \hline
    Kangaroo & 14862.5 & \textbf{14800} &(4.94)& 14500&(4.84) & 6740\\
    \hline
    Krull & 8627.9 & 10647 &(8.48)& \textbf{19962} &(17.2) & 3805\\
    \hline
    Kung Fu Master & 36733.3 & 62200 &(2.76) & \textbf{117700}&(5.22) & 23270\\
    \hline
  Montezuma Revenge & 100 & 0 &(0) & \textbf{2600}&(0.55) & 0\\
    \hline
  Ms Pacman & 2983.3 & 7250 &(1.04)& \textbf{13911} &(2.05) & 2311\\
    \hline
  Name This Game & 11501.1 & 14470 &(2.12)& \textbf{16820}&(2.52) & 7257\\
    \hline
  Pong & 20.9 & \textbf{21} &(1.18)& \textbf{21} &(1.18) & 18.9\\
    \hline
  Private Eye & 1812.5 & 200 & (0) & \textbf{14000} &(0.2) & 1788\\
    \hline
  Qbert & 15092.7 & 18475 &(1.38)& \textbf{32425} &(2.43) & 10596\\
    \hline
  Riverraid & 12845 & 18320 &(1.08)& \textbf{27790}&(1.68) & 8316\\
    \hline
  Road Runner & 51500 & 67500 &(8.62) & \textbf{69500}&(8.87) & 18257\\
    \hline
  Robotank & 66.6 & 62 &(6.16)& \textbf{81}&(8.12) & 51.6\\
    \hline
  Seaquest & 9083.1 & 30380 &(0.72)& \textbf{32840} &(0.78) & 5286\\
    \hline
   Space Invaders & 2893 & 2250 &(1.38)& \textbf{3150} &(1.97)& 1976\\ 
    \hline
   Star Gunner & 55725 & 54000 &(5.56)& \textbf{83400}&(8.63) &  57997\\
    \hline
   Tennis & 0 & \textbf{24}& (3.08) & \textbf{24} &(3.08)&  -2.5\\
  \hline
   Time Pilot & 9079.4 & 12300&(5.26) & \textbf{132100} &(77.37)& 5947 \\
    \hline
   Tutankham & 214.8 & 394 &(2.45)& \textbf{396} &(2.46)&  186.7\\
    \hline
   Up N Down & 26231 & 38600&(3.41) & \textbf{174820} &(15.62) & 8456\\
    \hline
   Venture & 212.5 & 1700&(1.43) & \textbf{2000} &(1.68) & 380\\
    \hline
   Video Pinball & 811610 & 970906 &(54.95)& \textbf{986525} &(55.84)& 42684 \\
    \hline
   Wizard of Wor & 6804.7 & 25100 &(5.85)& \textbf{59900}&(14.15) & 3393\\
    \hline
   Zaxxon & 11491.7 & 20300&(2.22) & \textbf{28300} & (3.09) & 4977\\
    \hline
\end{tabular}

\label{tbl:res1}
\end{table*}

\begin{table*}[t]
\scriptsize
\caption{Mean Evaluation Scores, Standard Deviations, Human Normalized Scores and AUCs}

\centering
\begin{tabular}{ |c||c c c c|c c c c|} 
 \hline
    \textbf{Game} & \textbf{Double DQN} & \textbf{Std.}&\textbf{Normalized}  & \textbf{AUC}&\textbf{Swin DQN} &\textbf{Std.} &\textbf{Normalized} & \textbf{AUC}\\
  \hline
Alien & 1664.0 & 739.96 & 0.21 &0.52 & \textbf{5324.0} & 3868.63 & 0.74 & 1.08\\
\hline
Amidar & \textbf{604.0} & 63.5 & 0.35 & 0.68& 513.6 & 247.56 & 0.3 & 1.2\\
\hline
Assault & 1712.2 & 345.37 & 2.87 & 0.47& \textbf{3798.8} & 2432.75 & 6.88 & 0.53\\
\hline
Asterix & 25220.0 & 11338.15 & 3.02 & 1.68 & \textbf{596300.0} & 253477.93 & 71.88 & 27.84\\
\hline
Asteroids & 1102.0 & 303.87 & 0.01 & 0.49& \textbf{5046.0} & 2509.91 & 0.09 & 1.14\\
\hline
Atlantis & 753880.0 & 1327.25 & 45.8 &7.08& \textbf{835940.0} & 20888.43 & 50.88 &7.51\\
\hline
Bank Heist & \textbf{696.0} & 63.75 & 0.92 & 1.71& 552.0 & 100.88 & 0.73 &1.62\\
\hline
Battle Zone & 27000.0 & 7099.3 & 0.71 & 0.87&\textbf{61600.0} & 13749.18 & 1.7 &1.48\\
\hline
Beam Rider & 7952.4 & 2875.24 & 0.46 & 0.86&\textbf{24140.0} & 10361.26 & 1.44 &2.2\\
\hline
Bowling & \textbf{30.0} & 0.0 & 0.05 &0.4& 19.8 & 18.27 & -0.02 &1.33\\
\hline
Boxing & 91.2 & 1.6 & 7.59 &1.23& \textbf{99.2} & 0.75 & 8.26 &1.27\\
\hline
Breakout & 184.2 & 159.12 & 6.34 &0.8& \textbf{857.8} & 6.97 & 29.73 &1.1\\
\hline
Centipede & 3933.0 & 1059.55 & 0.19 & 0.38&\textbf{7071.0} & 1382.07 & 0.5 &0.73\\
\hline
Chopper Command & 620.0 & 146.97 & -0.03 &0.17 & \textbf{2120.0} & 584.47 & 0.2 &0.22\\
\hline
Crazy Climber & 103320.0 & 15244.2 & 3.69 &0.94& \textbf{118160.0} & 5443.01 & 4.29 &1.12\\
\hline
Demon Attack & 5533.0 & 3037.95 & 2.96 &0.39& \textbf{89254.0} & 49893.56 & 48.99 &1.69\\
\hline
Double Dunk & \textbf{0.4} & 0.8 & 8.64 &-0.08& \textbf{0.4} & 1.5 & 8.64 & -0.12\\
\hline
Enduro & 1038.6 & 189.68 & 1.21 &3.06& \textbf{1901.4} & 461.78 & 2.21 &4.67\\
\hline
Fishing Derby & -43.4 & 20.41 & 0.91 &-51.59& \textbf{17.4} & 8.62 & 2.06 &-34.48\\
\hline
Freeway & 21.6 & 1.2 & 0.73 &0.69& \textbf{34.0} & 0.0 & 1.15 &1.09\\
\hline
Frostbite & 1002.0 & 929.83 & 0.22 & 3.77& \textbf{2036.0} & 1024.04 & 0.46 &2.26\\
\hline
Gopher & \textbf{21496.0} & 12332.76 & 9.86 &1.89& 12216.0 & 3511.93 & 5.55 &1.59\\
\hline
Gravitar & \textbf{210.0} & 139.28 & 0.01 &0.99& 70.0 & 140.0 & -0.03 &0.43\\
\hline
Hero & 12716.0 & 192.0 & 0.39 &0.38& \textbf{27819.0} & 176.42 & 0.9 &0.85\\
\hline
Ice Hockey & -8.6 & 2.58 & 0.21 &-6.1& \textbf{-0.6} & 0.8 & 0.88 &-2.27\\
\hline
Jamesbond & 680.0 & 97.98 & 2.38 &0.99& \textbf{1980.0} & 697.57 & 7.13 &2.55\\
\hline
Kangaroo & \textbf{8660.0} & 1539.61 & 2.89 &0.98& 8280.0 & 1729.05 & 2.76 &1.36\\
\hline
Krull & \textbf{8385.0} & 771.32 & 6.36 &1.76 & 6861.6 & 1060.26 & 4.93 &1.77\\
\hline
Kung Fu Master & 21300.0 & 4285.79 & 0.94 &1.15 & \textbf{70640.0} & 15710.46 & 3.13 &1.9 \\
\hline
Montezuma Revenge & 0.0 & 0.0 & 0.0 & -& \textbf{100.0} & 0.0 & 0.02 &-\\
\hline
Ms Pacman & 2408.0 & 811.75 & 0.32 &1.08& \textbf{5878.0} & 2426.4 & 0.84 &2.13\\
\hline
Name This Game & 9140.0 & 1193.05 & 1.19 &1.22& \textbf{9670.0} & 2263.6 & 1.28 &1.28\\
\hline
Pong & \textbf{21.0} & 0.0 & 1.18 &1.01& \textbf{21.0} & 0.0 & 1.18 &1.05\\
\hline
Private Eye & 0.0 & 0.0 & -0.0 &0.03& \textbf{100.0} & 0.0 & 0.0 &0.03\\
\hline
Qbert & 15195.0 & 185.34 & 1.13 &1.15& \textbf{25365.0} & 1660.17 & 1.9 &1.61\\
\hline
Riverraid & 8106.0 & 224.64 & 0.43 &1.11& \textbf{20180.0} & 5523.52 & 1.19 &1.37\\
\hline
Road Runner & 46560.0 & 6021.49 & 5.94 &2.72& \textbf{56900.0} & 6829.35 & 7.26 &2.97\\
\hline
Robotank & 40.8 & 10.26 & 3.98 &0.67& \textbf{62.6} & 7.55 & 6.23 &1.03\\
\hline
Seaquest & \textbf{7852.0} & 1442.97 & 0.19&1.05 & 7480.0 & 1617.65 & 0.18 &1.01\\
\hline
Space Invaders & 1041.0 & 379.5 & 0.59 &0.43& \textbf{2113.0} & 879.69 & 1.29 &0.54\\
\hline
Star Gunner & 44980.0 & 8585.66 & 4.62 &0.36& \textbf{58480.0} & 11361.94 & 6.03 &0.65\\
\hline
Tennis & 20.6 & 1.02 & 2.86 &1.66& \textbf{23.8} & 0.4 & 3.07 &5.77 \\
\hline
Time Pilot & 9100.0 & 3919.69 & 3.33&0.72 & \textbf{51920.0} & 20264.79 & 29.11 &4.2\\
\hline
Tutankham & 216.8 & 64.39 & 1.31 &1.04& \textbf{251.6} & 42.32 & 1.54 &1.21\\
\hline
Up N Down & 7876.0 & 4448.75 & 0.66 &1.11& \textbf{34650.0} & 8173.73 & 3.06 &1.75\\
\hline
Venture & \textbf{940.0} & 120.0 & 0.79 &1.83 & 500.0 & 154.92 & 0.42 &1.43\\
\hline
Video Pinball & 207074.6 & 201917.13 & 11.72 &2.62& \textbf{256106.4} & 179668.88 & 14.5 &2.05\\
\hline
Wizard Of Wor & 1900.0 & 252.98 & 0.32 &1.1& \textbf{14700.0} & 6867.02 & 3.37 &3.38\\
\hline
Zaxxon & 7020.0 & 406.94 & 0.76 &1.48& \textbf{13820.0} & 936.8 & 1.51 &2.1\\
\hline
\end{tabular}
\label{tbl:res2}
\end{table*}

\begin{figure*}[t]
    \centering
    \includegraphics[width=0.8\linewidth]{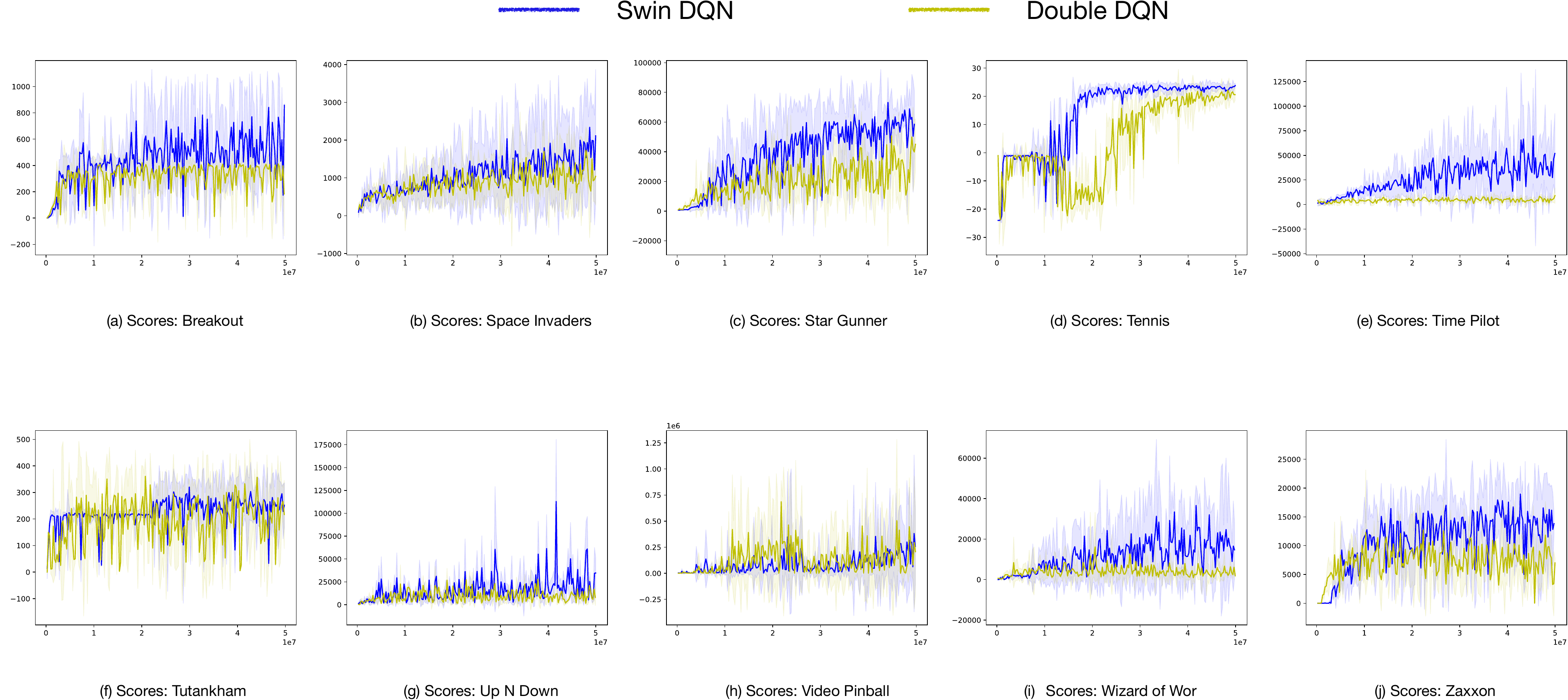}
    \caption{The mean evaluation scores together with 95\% confidence intervals during 200M frames (50M training steps). The blue line is Swin DQN. The orange line is Double DQN. The interval between every evaluation is 1M frames (250000 steps).}
\label{fig:res1}
\end{figure*}

In this section, we consider the mean evaluation scores, together with the evaluation curves for a more detailed examination. Complete results over 49 games are included in Table \ref{tbl:res1} and Table \ref{tbl:res2}. The mean evaluation scores and confidence intervals for 10 games are shown in Fig. \ref{fig:res1}. Each evaluation score is a mean of 5 runs from randomly seeded environments in a single training round.

Table \ref{tbl:res1} contains the maximal evaluation scores, whereas Table \ref{tbl:res2} contains the full mean evaluation scores, standard deviations, human normalized scores and area under curve (AUC) \cite{stadie2015incentivizing}. AUCs are the cumulative evaluation scores normalized by the final scores from \cite{mnih2015human} over the entire training period.  Algorithms with the best performance for each game are highlighted. The key results are the comparisons between Swin DQN and Double DQN, since they have no algorithmic difference and the only change is that Swin DQN utilizes the Swin backbone instead of CNNs.

In Table \ref{tbl:res1}, we compare Swin DQN, Double DQN, Bootstrapped DQN (Boot-DQN) \cite{osband2016deep} and the Nature DQN \cite{mnih2015human} because they share the most similarities in experimental design and hyper-parameter choices. It is clear that Swin DQN performs the best out of the four algorithms in terms of maximal evaluation scores. The highest evaluation scores of Swin DQN are higher or equal than those of Double DQN in 45 out of 49 games (91.84\%), better than the mean evaluation scores. Its human normalized scores also suggest that our algorithm is capable of human or even superhuman level performances in most of the games. The mean evaluation scores of Swin DQN are higher or equal than those of Double DQN in 40 out of 49 games (81.63\%). AUCs also entail that the gap between their score curves remain at a significant level for most of the training process for each game.




In many games, the improvements are as significant as multiple times the evaluation scores of Double DQN, but for few games, the improvement is less prominent. This discrepancy may stem from the variety of Atari games. Different games have various unique settings and require distinguished sets of skills. Entity/pixel attention connections are crucial in order to solve puzzles in some games, but they might contribute trivially to other games. Swin Transformers appear to have a forceful impact in games that have a high complexity of features or require subtle world modeling.

Swin DQN also shows drastically improved scores over Double DQN throughout the training process except for the beginning in Fig. \ref{fig:res1}. For instance, we can see that there is a significant difference of scores for those two models from 1.5M to 3M steps in Tennis from Fig. \ref{fig:res1}. Swin DQN converges earlier at around 2M steps and Double DQN at around 4M steps. Double DQN being stuck in suboptimal performances longer indicates a more robust modeling capability of Swin Transformers on adversity. In most of the games, Swin DQN does not show faster converging speed than Double DQN in the initial training frames. This observation corresponds to the higher complexity of the Swin model. The Swin transformers designed in our experiment have a larger network structure and, in turn, requires more input data than CNNs. Thus, the strength of Swin DQN is demonstrated only after a proper amount of training updates.

\begin{figure}[t]
    \centering
    \includegraphics[width=0.3\linewidth]{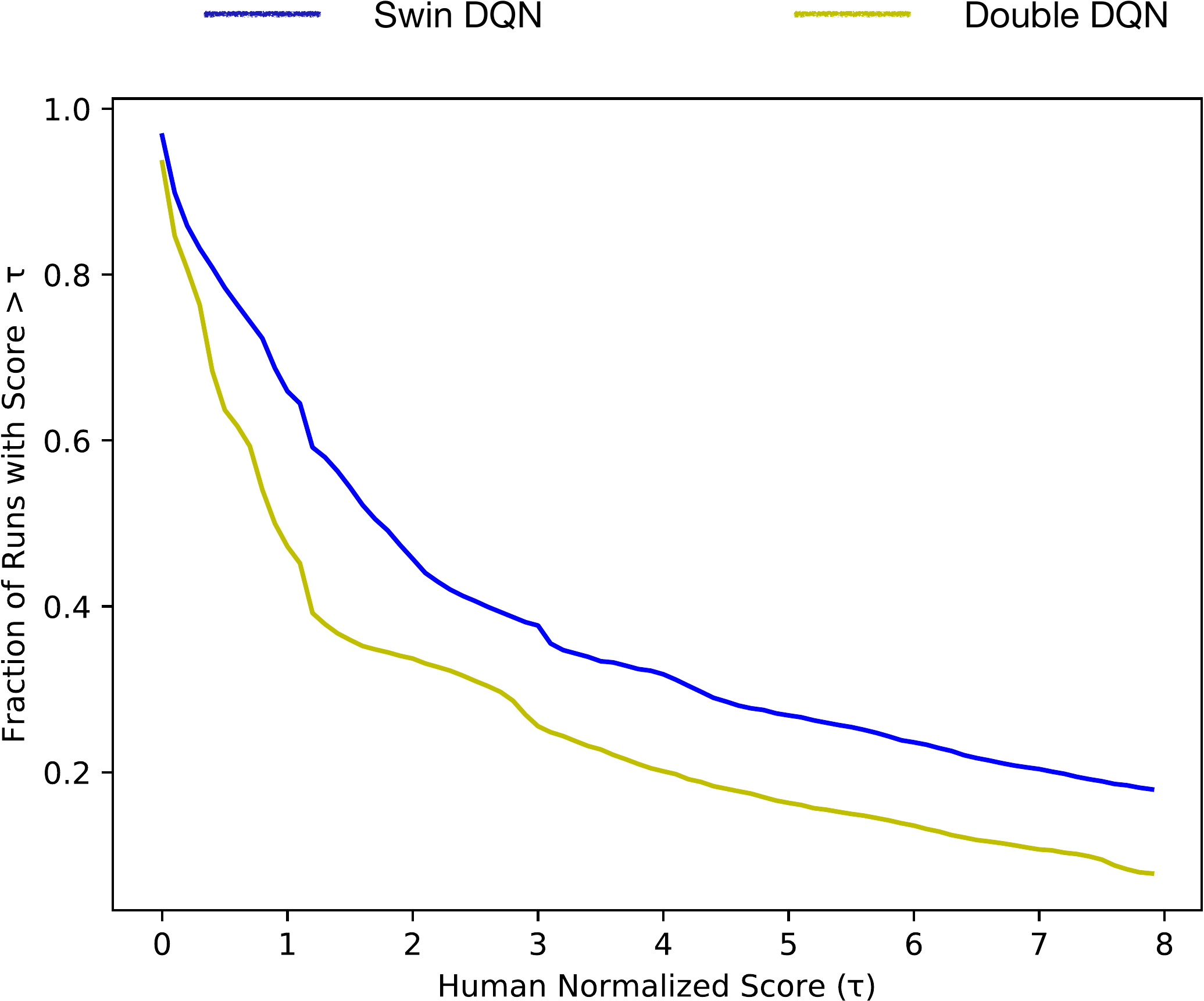}
    \caption{The performance profiles up to $\tau=8$ that shows the percentage of the score larger than $\tau$. The blue line is Swin DQN and the orange line is Double DQN.}
\label{fig:pp}
\end{figure}

We also include performance profiles (score distributions) described in \cite{agarwal2021deep} with overall human normalized scores in Fig. \ref{fig:pp}, which allow an aggregated qualitative comparison between Swin DQN and Double DQN. The gap between two lines is non-trivial as $\tau$ increases, entailing that the improvement of human normalized scores among 49 games is salient when replacing CNNs with Swin Transformers.

\begin{figure*}[t]
    \centering
    \includegraphics[width=0.8\linewidth]{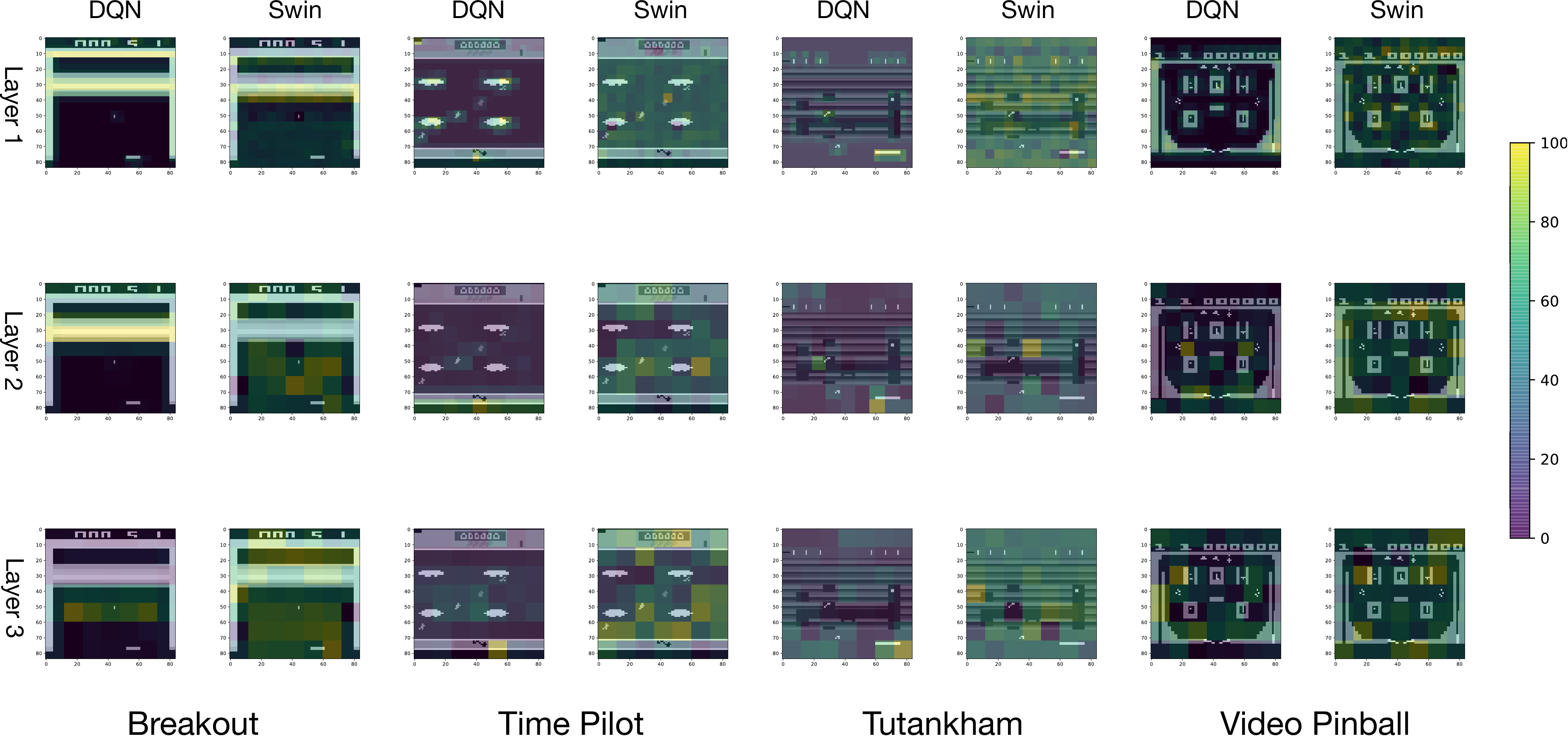}
    \caption{The activation maps of our Double DQN and Swin DQN. All models are trained for 200M frames. The image background is in gray scale. As the color bar suggests, the more yellow the pixel gets, the higher the activation.}
\label{fig:ac}
\end{figure*}

In a few of the games (e.g., Video Pinball, Tutankham), Swin DQN shows tiny improvements compared to Double DQN during the training process, but still reaches slightly higher scores than Double DQN at the end. It is typical that Double DQN performs reasonably well in those games. Thus, the advantages of better feature representations through introducing Swin Transformers are relatively obscure. To further examine this observation, activation maps of images from four different games fed into trained models are illustrated in Fig. \ref{fig:ac}. Among the four games, Breakout and Time Pilot are representative games in which Swin DQN shows significant improvements, while Video Pinball and Tutankham are games in which the improvements are relatively small.

As expected, Swin DQN possesses significantly richer representations than Double DQN especially in layer-2 and layer-3 activation maps from Fig. \ref{fig:ac}. Moreover, Swin DQN have great advantage when representing entity/pixel relationships, while Double DQN is only capable of reasonably capturing part of the scoreboard or stationary objects. In Breakout and Time Pilot, Swin DQN successfully captures some possible moving trajectories of the target objects, but Double DQN is not excelled at this, or constantly activates wrong trajectories. 

\section{Discussion}

We have successfully demonstrated that Swin Transformers can improve the performances of DQN significantly in Atari games. Spatial self-attentions benefit not only CV tasks but also image-based RL tasks, and introducing spatial self-attentions alone is sufficient to improve the performances of current DRL models drastically. The structure of Swin Transformers also makes local attentions more affordable in DRL.


Our hyper-parameters of the Swin Transformer backbone are chosen by grid search, yet its structure might not be with the least necessary training parameters. Swin DQN is still computationally more costly (3 to 4 times in wall clock time) than DQN. It is appealing to seek for the smallest Swin model without jeopardizing its performances. Thus, experimenting Swin DQN with less training parameters can be crucial future work.

The positive effects of using self-attentions in Atari games lead us to the expectation that real-world control tasks can also exploit self-attentions. We humans often need to direct our attention to a small area in our visual filed when solving physical control tasks, and this leaves room for transformers. DRL can be potentially more applicable in solving physical problems with the help of Swin Transformers in the future.

 \begin{acks}
 This work was performed on the [ML node] resource, owned by the University of Oslo, and operated by the Department for Research Computing at USIT, the University of Oslo IT-department.
\end{acks}
\bibliographystyle{ACM-Reference-Format}
\bibliography{main}


\begin{thebibliography}{31}


\ifx \showCODEN    \undefined \def \showCODEN     #1{\unskip}     \fi
\ifx \showDOI      \undefined \def \showDOI       #1{#1}\fi
\ifx \showISBNx    \undefined \def \showISBNx     #1{\unskip}     \fi
\ifx \showISBNxiii \undefined \def \showISBNxiii  #1{\unskip}     \fi
\ifx \showISSN     \undefined \def \showISSN      #1{\unskip}     \fi
\ifx \showLCCN     \undefined \def \showLCCN      #1{\unskip}     \fi
\ifx \shownote     \undefined \def \shownote      #1{#1}          \fi
\ifx \showarticletitle \undefined \def \showarticletitle #1{#1}   \fi
\ifx \showURL      \undefined \def \showURL       {\relax}        \fi
\providecommand\bibfield[2]{#2}
\providecommand\bibinfo[2]{#2}
\providecommand\natexlab[1]{#1}
\providecommand\showeprint[2][]{arXiv:#2}

\bibitem[\protect\citeauthoryear{Agarwal, Schwarzer, Castro, Courville, and
  Bellemare}{Agarwal et~al\mbox{.}}{2021}]%
        {agarwal2021deep}
\bibfield{author}{\bibinfo{person}{Rishabh Agarwal}, \bibinfo{person}{Max
  Schwarzer}, \bibinfo{person}{Pablo~Samuel Castro}, \bibinfo{person}{Aaron~C
  Courville}, {and} \bibinfo{person}{Marc Bellemare}.}
  \bibinfo{year}{2021}\natexlab{}.
\newblock \showarticletitle{Deep reinforcement learning at the edge of the
  statistical precipice}.
\newblock \bibinfo{journal}{\emph{Advances in Neural Information Processing
  Systems}}  \bibinfo{volume}{34} (\bibinfo{year}{2021}).
\newblock


\bibitem[\protect\citeauthoryear{Caron, Touvron, Misra, J{\'e}gou, Mairal,
  Bojanowski, and Joulin}{Caron et~al\mbox{.}}{2021}]%
        {caron2021emerging}
\bibfield{author}{\bibinfo{person}{Mathilde Caron}, \bibinfo{person}{Hugo
  Touvron}, \bibinfo{person}{Ishan Misra}, \bibinfo{person}{Herv{\'e}
  J{\'e}gou}, \bibinfo{person}{Julien Mairal}, \bibinfo{person}{Piotr
  Bojanowski}, {and} \bibinfo{person}{Armand Joulin}.}
  \bibinfo{year}{2021}\natexlab{}.
\newblock \showarticletitle{Emerging properties in self-supervised vision
  transformers}. In \bibinfo{booktitle}{\emph{Proceedings of the IEEE/CVF
  International Conference on Computer Vision}}. \bibinfo{pages}{9650--9660}.
\newblock


\bibitem[\protect\citeauthoryear{Chen, Lu, Rajeswaran, Lee, Grover, Laskin,
  Abbeel, Srinivas, and Mordatch}{Chen et~al\mbox{.}}{2021a}]%
        {chen2021decision}
\bibfield{author}{\bibinfo{person}{Lili Chen}, \bibinfo{person}{Kevin Lu},
  \bibinfo{person}{Aravind Rajeswaran}, \bibinfo{person}{Kimin Lee},
  \bibinfo{person}{Aditya Grover}, \bibinfo{person}{Misha Laskin},
  \bibinfo{person}{Pieter Abbeel}, \bibinfo{person}{Aravind Srinivas}, {and}
  \bibinfo{person}{Igor Mordatch}.} \bibinfo{year}{2021}\natexlab{a}.
\newblock \showarticletitle{Decision transformer: Reinforcement learning via
  sequence modeling}.
\newblock \bibinfo{journal}{\emph{Advances in neural information processing
  systems}}  \bibinfo{volume}{34} (\bibinfo{year}{2021}).
\newblock


\bibitem[\protect\citeauthoryear{Chen, Xie, and He}{Chen
  et~al\mbox{.}}{2021b}]%
        {chen2021empirical}
\bibfield{author}{\bibinfo{person}{Xinlei Chen}, \bibinfo{person}{Saining Xie},
  {and} \bibinfo{person}{Kaiming He}.} \bibinfo{year}{2021}\natexlab{b}.
\newblock \showarticletitle{An empirical study of training self-supervised
  vision transformers}. In \bibinfo{booktitle}{\emph{Proceedings of the
  IEEE/CVF International Conference on Computer Vision}}.
  \bibinfo{pages}{9640--9649}.
\newblock
\newblock
\shownote{10.1109/ICCV48922.2021.00950.}


\bibitem[\protect\citeauthoryear{Dosovitskiy, Beyer, Kolesnikov, Weissenborn,
  Zhai, Unterthiner, Dehghani, Minderer, Heigold, Gelly,
  et~al\mbox{.}}{Dosovitskiy et~al\mbox{.}}{2021}]%
        {dosovitskiy2020image}
\bibfield{author}{\bibinfo{person}{Alexey Dosovitskiy}, \bibinfo{person}{Lucas
  Beyer}, \bibinfo{person}{Alexander Kolesnikov}, \bibinfo{person}{Dirk
  Weissenborn}, \bibinfo{person}{Xiaohua Zhai}, \bibinfo{person}{Thomas
  Unterthiner}, \bibinfo{person}{Mostafa Dehghani}, \bibinfo{person}{Matthias
  Minderer}, \bibinfo{person}{Georg Heigold}, \bibinfo{person}{Sylvain Gelly},
  {et~al\mbox{.}}} \bibinfo{year}{2021}\natexlab{}.
\newblock \showarticletitle{An image is worth 16x16 words: Transformers for
  image recognition at scale}.
\newblock \bibinfo{journal}{\emph{ICLR}} (\bibinfo{year}{2021}).
\newblock


\bibitem[\protect\citeauthoryear{Hasselt}{Hasselt}{2010}]%
        {hasselt2010double}
\bibfield{author}{\bibinfo{person}{Hado Hasselt}.}
  \bibinfo{year}{2010}\natexlab{}.
\newblock \showarticletitle{Double Q-learning}. In
  \bibinfo{booktitle}{\emph{Advances in neural information processing systems
  23}}. \bibinfo{publisher}{Curran Associates, Inc.}, \bibinfo{address}{Red
  Hook, NY}, \bibinfo{pages}{2613--2621}.
\newblock


\bibitem[\protect\citeauthoryear{He, Fan, Wu, Xie, and Girshick}{He
  et~al\mbox{.}}{2020}]%
        {he2020momentum}
\bibfield{author}{\bibinfo{person}{Kaiming He}, \bibinfo{person}{Haoqi Fan},
  \bibinfo{person}{Yuxin Wu}, \bibinfo{person}{Saining Xie}, {and}
  \bibinfo{person}{Ross Girshick}.} \bibinfo{year}{2020}\natexlab{}.
\newblock \showarticletitle{Momentum contrast for unsupervised visual
  representation learning}. In \bibinfo{booktitle}{\emph{Proceedings of the
  IEEE/CVF conference on computer vision and pattern recognition}}.
  \bibinfo{pages}{9729--9738}.
\newblock
\newblock
\shownote{10.1109/cvpr42600.2020.00975.}


\bibitem[\protect\citeauthoryear{Kalantari, Amini, Chandar, and
  Precup}{Kalantari et~al\mbox{.}}{2022}]%
        {kalantari2022improving}
\bibfield{author}{\bibinfo{person}{Amir~Ardalan Kalantari},
  \bibinfo{person}{Mohammad Amini}, \bibinfo{person}{Sarath Chandar}, {and}
  \bibinfo{person}{Doina Precup}.} \bibinfo{year}{2022}\natexlab{}.
\newblock \showarticletitle{Improving Sample Efficiency of Value Based Models
  Using Attention and Vision Transformers}.
\newblock \bibinfo{journal}{\emph{arXiv preprint arXiv:2202.00710}}
  (\bibinfo{year}{2022}).
\newblock
\newblock
\shownote{10.48550/arXiv.2202.00710.}


\bibitem[\protect\citeauthoryear{Kargar and Kyrki}{Kargar and Kyrki}{2021}]%
        {kargar2021vision}
\bibfield{author}{\bibinfo{person}{Eshagh Kargar} {and} \bibinfo{person}{Ville
  Kyrki}.} \bibinfo{year}{2021}\natexlab{}.
\newblock \showarticletitle{Vision transformer for learning driving policies in
  complex multi-agent environments}.
\newblock \bibinfo{journal}{\emph{arXiv preprint}} (\bibinfo{year}{2021}).
\newblock


\bibitem[\protect\citeauthoryear{Kingma and Ba}{Kingma and Ba}{2014}]%
        {kingma2014adam}
\bibfield{author}{\bibinfo{person}{Diederik~P Kingma} {and}
  \bibinfo{person}{Jimmy Ba}.} \bibinfo{year}{2014}\natexlab{}.
\newblock \showarticletitle{Adam: A method for stochastic optimization}.
\newblock \bibinfo{journal}{\emph{arXiv preprint arXiv:1412.6980}}
  (\bibinfo{year}{2014}).
\newblock
\newblock
\shownote{10.48550/arXiv.1412.6980.}


\bibitem[\protect\citeauthoryear{Li, Li, Huang, and Yu}{Li
  et~al\mbox{.}}{2022}]%
        {li2022interactive}
\bibfield{author}{\bibinfo{person}{Changyang Li}, \bibinfo{person}{Wanwan Li},
  \bibinfo{person}{Haikun Huang}, {and} \bibinfo{person}{Lap-Fai Yu}.}
  \bibinfo{year}{2022}\natexlab{}.
\newblock \showarticletitle{Interactive augmented reality storytelling guided
  by scene semantics}.
\newblock \bibinfo{journal}{\emph{ACM Transactions on Graphics (TOG)}}
  \bibinfo{volume}{41}, \bibinfo{number}{4} (\bibinfo{year}{2022}),
  \bibinfo{pages}{1--15}.
\newblock


\bibitem[\protect\citeauthoryear{Li}{Li}{2021}]%
        {li2021image}
\bibfield{author}{\bibinfo{person}{Wanwan Li}.}
  \bibinfo{year}{2021}\natexlab{}.
\newblock \showarticletitle{Image Synthesis and Editing with Generative
  Adversarial Networks (GANs): A Review}. In \bibinfo{booktitle}{\emph{2021
  Fifth World Conference on Smart Trends in Systems Security and Sustainability
  (WorldS4)}}. IEEE, \bibinfo{pages}{65--70}.
\newblock


\bibitem[\protect\citeauthoryear{Li}{Li}{2022}]%
        {li2022musical}
\bibfield{author}{\bibinfo{person}{Wanwan Li}.}
  \bibinfo{year}{2022}\natexlab{}.
\newblock \showarticletitle{Musical Instrument Performance in Augmented
  Virtuality}. In \bibinfo{booktitle}{\emph{Proceedings of the 6th
  International Conference on Digital Signal Processing}}.
  \bibinfo{pages}{91--97}.
\newblock


\bibitem[\protect\citeauthoryear{Li}{Li}{2023a}]%
        {li2023synthesizing}
\bibfield{author}{\bibinfo{person}{Wanwan Li}.}
  \bibinfo{year}{2023}\natexlab{a}.
\newblock \showarticletitle{Synthesizing 3D VR Sketch Using Generative
  Adversarial Neural Network}. In \bibinfo{booktitle}{\emph{Proceedings of the
  2023 7th International Conference on Big Data and Internet of Things}}.
  \bibinfo{pages}{122--128}.
\newblock


\bibitem[\protect\citeauthoryear{Li}{Li}{2023b}]%
        {li2023terrain}
\bibfield{author}{\bibinfo{person}{Wanwan Li}.}
  \bibinfo{year}{2023}\natexlab{b}.
\newblock \showarticletitle{Terrain synthesis for treadmill exergaming in
  virtual reality}. In \bibinfo{booktitle}{\emph{2023 IEEE Conference on
  Virtual Reality and 3D User Interfaces Abstracts and Workshops (VRW)}}. IEEE,
  \bibinfo{pages}{263--269}.
\newblock


\bibitem[\protect\citeauthoryear{Li, Li, Kim, Huang, and Yu}{Li
  et~al\mbox{.}}{2023}]%
        {li2023location}
\bibfield{author}{\bibinfo{person}{Wanwan Li}, \bibinfo{person}{Changyang Li},
  \bibinfo{person}{Minyoung Kim}, \bibinfo{person}{Haikun Huang}, {and}
  \bibinfo{person}{Lap-Fai Yu}.} \bibinfo{year}{2023}\natexlab{}.
\newblock \showarticletitle{Location-Aware Adaptation of Augmented Reality
  Narratives}. In \bibinfo{booktitle}{\emph{Proceedings of the 2023 CHI
  Conference on Human Factors in Computing Systems}}. \bibinfo{pages}{1--15}.
\newblock


\bibitem[\protect\citeauthoryear{Li, Xie, Zhang, Meiss, Huang, and Yu}{Li
  et~al\mbox{.}}{2020}]%
        {li2020exertion}
\bibfield{author}{\bibinfo{person}{Wanwan Li}, \bibinfo{person}{Biao Xie},
  \bibinfo{person}{Yongqi Zhang}, \bibinfo{person}{Walter Meiss},
  \bibinfo{person}{Haikun Huang}, {and} \bibinfo{person}{Lap-Fai Yu}.}
  \bibinfo{year}{2020}\natexlab{}.
\newblock \showarticletitle{Exertion-aware path generation}.
\newblock \bibinfo{journal}{\emph{ACM Trans. Graph.}} \bibinfo{volume}{39},
  \bibinfo{number}{4} (\bibinfo{year}{2020}), \bibinfo{pages}{115}.
\newblock


\bibitem[\protect\citeauthoryear{Liu, Lin, Cao, Hu, Wei, Zhang, Lin, and
  Guo}{Liu et~al\mbox{.}}{2021}]%
        {liu2021swin}
\bibfield{author}{\bibinfo{person}{Ze Liu}, \bibinfo{person}{Yutong Lin},
  \bibinfo{person}{Yue Cao}, \bibinfo{person}{Han Hu}, \bibinfo{person}{Yixuan
  Wei}, \bibinfo{person}{Zheng Zhang}, \bibinfo{person}{Stephen Lin}, {and}
  \bibinfo{person}{Baining Guo}.} \bibinfo{year}{2021}\natexlab{}.
\newblock \showarticletitle{Swin transformer: Hierarchical vision transformer
  using shifted windows}. In \bibinfo{booktitle}{\emph{Proceedings of the
  IEEE/CVF International Conference on Computer Vision}}.
  \bibinfo{pages}{10012--10022}.
\newblock


\bibitem[\protect\citeauthoryear{Meng, Yazidi, Goodwin, and Engelstad}{Meng
  et~al\mbox{.}}{2021}]%
        {meng2021expert}
\bibfield{author}{\bibinfo{person}{Li Meng}, \bibinfo{person}{Anis Yazidi},
  \bibinfo{person}{Morten Goodwin}, {and} \bibinfo{person}{Paal Engelstad}.}
  \bibinfo{year}{2021}\natexlab{}.
\newblock \showarticletitle{Expert Q-learning: Deep Q-learning With State
  Values From Expert Examples.}
\newblock \bibinfo{journal}{\emph{CoRR}} (\bibinfo{year}{2021}).
\newblock
\newblock
\shownote{10.7557/18.6237.}


\bibitem[\protect\citeauthoryear{Mnih, Kavukcuoglu, Silver, Graves, Antonoglou,
  Wierstra, and Riedmiller}{Mnih et~al\mbox{.}}{2013}]%
        {mnih2013playing}
\bibfield{author}{\bibinfo{person}{Volodymyr Mnih}, \bibinfo{person}{Koray
  Kavukcuoglu}, \bibinfo{person}{David Silver}, \bibinfo{person}{Alex Graves},
  \bibinfo{person}{Ioannis Antonoglou}, \bibinfo{person}{Daan Wierstra}, {and}
  \bibinfo{person}{Martin Riedmiller}.} \bibinfo{year}{2013}\natexlab{}.
\newblock \bibinfo{title}{Playing Atari with Deep Reinforcement Learning}.
\newblock
\newblock
\showeprint[arxiv]{1312.5602}~[cs.LG]
\newblock
\shownote{10.48550/arXiv.1312.5602.}


\bibitem[\protect\citeauthoryear{Mnih, Kavukcuoglu, Silver, Rusu, Veness,
  Bellemare, Graves, Riedmiller, Fidjeland, Ostrovski, et~al\mbox{.}}{Mnih
  et~al\mbox{.}}{2015}]%
        {mnih2015human}
\bibfield{author}{\bibinfo{person}{Volodymyr Mnih}, \bibinfo{person}{Koray
  Kavukcuoglu}, \bibinfo{person}{David Silver}, \bibinfo{person}{Andrei~A
  Rusu}, \bibinfo{person}{Joel Veness}, \bibinfo{person}{Marc~G Bellemare},
  \bibinfo{person}{Alex Graves}, \bibinfo{person}{Martin Riedmiller},
  \bibinfo{person}{Andreas~K Fidjeland}, \bibinfo{person}{Georg Ostrovski},
  {et~al\mbox{.}}} \bibinfo{year}{2015}\natexlab{}.
\newblock \showarticletitle{Human-level control through deep reinforcement
  learning}.
\newblock \bibinfo{journal}{\emph{nature}} \bibinfo{volume}{518},
  \bibinfo{number}{7540} (\bibinfo{year}{2015}), \bibinfo{pages}{529--533}.
\newblock


\bibitem[\protect\citeauthoryear{Osband, Blundell, Pritzel, and Van~Roy}{Osband
  et~al\mbox{.}}{2016}]%
        {osband2016deep}
\bibfield{author}{\bibinfo{person}{Ian Osband}, \bibinfo{person}{Charles
  Blundell}, \bibinfo{person}{Alexander Pritzel}, {and}
  \bibinfo{person}{Benjamin Van~Roy}.} \bibinfo{year}{2016}\natexlab{}.
\newblock \showarticletitle{Deep exploration via bootstrapped DQN}.
\newblock \bibinfo{journal}{\emph{Advances in neural information processing
  systems}}  \bibinfo{volume}{29} (\bibinfo{year}{2016}),
  \bibinfo{pages}{4026--4034}.
\newblock
\urldef\tempurl%
\url{https://doi.org/10.5555/3157382.3157548}
\showDOI{\tempurl}


\bibitem[\protect\citeauthoryear{S.~Gao and W}{S.~Gao and W}{2024}]%
        {gao2024image}
\bibfield{author}{\bibinfo{person}{B.~Hui S.~Gao} {and} \bibinfo{person}{Li
  W}.} \bibinfo{year}{2024}\natexlab{}.
\newblock \showarticletitle{Image Generation of Egyptian Hieroglyphs}. In
  \bibinfo{booktitle}{\emph{ICMLC 2024}}.
\newblock


\bibitem[\protect\citeauthoryear{Stadie, Levine, and Abbeel}{Stadie
  et~al\mbox{.}}{2015}]%
        {stadie2015incentivizing}
\bibfield{author}{\bibinfo{person}{Bradly~C Stadie}, \bibinfo{person}{Sergey
  Levine}, {and} \bibinfo{person}{Pieter Abbeel}.}
  \bibinfo{year}{2015}\natexlab{}.
\newblock \showarticletitle{Incentivizing exploration in reinforcement learning
  with deep predictive models}.
\newblock \bibinfo{journal}{\emph{arXiv preprint arXiv:1507.00814}}
  (\bibinfo{year}{2015}).
\newblock
\newblock
\shownote{10.48550/arXiv.1507.00814.}


\bibitem[\protect\citeauthoryear{Tao, Reda, and van~de Panne}{Tao
  et~al\mbox{.}}{2022}]%
        {tao2022evaluating}
\bibfield{author}{\bibinfo{person}{Tianxin Tao}, \bibinfo{person}{Daniele
  Reda}, {and} \bibinfo{person}{Michiel van~de Panne}.}
  \bibinfo{year}{2022}\natexlab{}.
\newblock \showarticletitle{Evaluating Vision Transformer Methods for Deep
  Reinforcement Learning from Pixels}.
\newblock \bibinfo{journal}{\emph{arXiv preprint arXiv:2204.04905}}
  (\bibinfo{year}{2022}).
\newblock
\newblock
\shownote{10.48550/arXiv.2204.04905.}


\bibitem[\protect\citeauthoryear{Thrun and Schwartz}{Thrun and
  Schwartz}{1993}]%
        {thrun1993issues}
\bibfield{author}{\bibinfo{person}{Sebastian Thrun} {and}
  \bibinfo{person}{Anton Schwartz}.} \bibinfo{year}{1993}\natexlab{}.
\newblock \showarticletitle{Issues in using function approximation for
  reinforcement learning}. In \bibinfo{booktitle}{\emph{Proceedings of the 4th
  Connectionist Models Summer School}}. \bibinfo{publisher}{Hillsdale},
  \bibinfo{address}{NJ}, \bibinfo{pages}{255--263}.
\newblock


\bibitem[\protect\citeauthoryear{Vaswani, Shazeer, Parmar, Uszkoreit, Jones,
  Gomez, Kaiser, and Polosukhin}{Vaswani et~al\mbox{.}}{2017}]%
        {vaswani2017attention}
\bibfield{author}{\bibinfo{person}{Ashish Vaswani}, \bibinfo{person}{Noam
  Shazeer}, \bibinfo{person}{Niki Parmar}, \bibinfo{person}{Jakob Uszkoreit},
  \bibinfo{person}{Llion Jones}, \bibinfo{person}{Aidan~N Gomez},
  \bibinfo{person}{{\L}ukasz Kaiser}, {and} \bibinfo{person}{Illia
  Polosukhin}.} \bibinfo{year}{2017}\natexlab{}.
\newblock \showarticletitle{Attention is all you need}.
\newblock \bibinfo{journal}{\emph{Advances in neural information processing
  systems}}  \bibinfo{volume}{30} (\bibinfo{year}{2017}).
\newblock
\newblock
\shownote{10.5555/3295222.3295349.}


\bibitem[\protect\citeauthoryear{Wang, Li, Khabsa, Fang, and Ma}{Wang
  et~al\mbox{.}}{2020}]%
        {wang2020self}
\bibfield{author}{\bibinfo{person}{Sinong Wang}, \bibinfo{person}{Belinda Li},
  \bibinfo{person}{Madian Khabsa}, \bibinfo{person}{Han Fang}, {and}
  \bibinfo{person}{H~Linformer Ma}.} \bibinfo{year}{2020}\natexlab{}.
\newblock \showarticletitle{Self-attention with linear complexity}.
\newblock \bibinfo{journal}{\emph{arXiv preprint}}  \bibinfo{volume}{8}
  (\bibinfo{year}{2020}).
\newblock


\bibitem[\protect\citeauthoryear{Watkins and Dayan}{Watkins and Dayan}{1992}]%
        {watkins1992q}
\bibfield{author}{\bibinfo{person}{Christopher~JCH Watkins} {and}
  \bibinfo{person}{Peter Dayan}.} \bibinfo{year}{1992}\natexlab{}.
\newblock \showarticletitle{Q-learning}.
\newblock \bibinfo{journal}{\emph{Machine learning}} \bibinfo{volume}{8},
  \bibinfo{number}{3-4} (\bibinfo{year}{1992}), \bibinfo{pages}{279--292}.
\newblock
\newblock
\shownote{10.1007/BF00992698.}


\bibitem[\protect\citeauthoryear{Y., W., W., D., and L.F}{Y.
  et~al\mbox{.}}{2020}]%
        {li2023sceneaware}
\bibfield{author}{\bibinfo{person}{Wang Y.}, \bibinfo{person}{Liang W.},
  \bibinfo{person}{Li W.}, \bibinfo{person}{Li D.}, {and} \bibinfo{person}{Yu
  L.F}.} \bibinfo{year}{2020}\natexlab{}.
\newblock \showarticletitle{Scene-aware background music synthesis}. In
  \bibinfo{booktitle}{\emph{Proceedings of the 28th ACM International
  Conference on Multimedia}}. \bibinfo{pages}{1162--1170}.
\newblock


\bibitem[\protect\citeauthoryear{Zheng, Zhang, and Grover}{Zheng
  et~al\mbox{.}}{2022}]%
        {zheng2022online}
\bibfield{author}{\bibinfo{person}{Qinqing Zheng}, \bibinfo{person}{Amy Zhang},
  {and} \bibinfo{person}{Aditya Grover}.} \bibinfo{year}{2022}\natexlab{}.
\newblock \showarticletitle{Online decision transformer}. In
  \bibinfo{booktitle}{\emph{International Conference on Machine Learning}}.
  PMLR, \bibinfo{pages}{27042--27059}.
\newblock


\end{thebibliography}

\end{document}